\documentclass[runningheads]{llncs}
\usepackage{graphicx}
\usepackage{xcolor}
\usepackage[caption=false]{subfig}

\usepackage{amssymb,amsmath,amsfonts}
\usepackage{bm}
\usepackage{multirow}
\usepackage[normalem]{ulem}

\usepackage{xcolor,colortbl}

\usepackage{hyperref}

\begin{document}
\title{Split and Expand: An inference-time improvement for Weakly Supervised Cell Instance Segmentation} %\thanks{Supported by organization x.}
\titlerunning{Split and Expand}
\author{Lin Geng Foo \inst{1} \and
Rui En Ho \inst{1} \and 
Jiamei Sun \inst{1} \and
Alexander Binder \inst{2,3} }
% %
\authorrunning{Foo et al.}
\institute{Singapore University of Technology and Design, Singapore \and
University of Oslo, Norway \and
Singapore Institute of Technology, Singapore
}
% % \email{lncs@springer.com}\\
% % \url{http://www.springer.com/gp/computer-science/lncs} \and
% % ABC Institute, Rupert-Karls-University Heidelberg, Heidelberg, Germany\\
\maketitle              % typeset the header of the contribution
\begin{abstract}
% The abstract should briefly summarize the contents of the paper in
% 15--250 words.
% We consider the problem of segmenting cell nuclei instances from Hematoxylin and Eosin (H\&E) stains with dot annotations only. While most recent works focus on improving the segmentation quality, this is usually insufficient for instance segmentation of cell instances clustered together or with a small size. In this work, we propose a simple two-step post-processing procedure, Split and Expand, that directly improves the conversion of segmentation maps to instances. In the splitting step, we generate fine-grained cell instances from the segmentation map with the guidance of cell-center predictions. For the expansion step, we utilize Layer-wise Relevance Propagation (LRP) explanation results to add small cells that are not captured in the segmentation map. Although we additionally train an output head to predict cell-centers, the post-processing procedure itself is not explicitly trained and is executed at inference-time only. A feature re-weighting loss based on LRP is proposed to improve our method even further. We test our procedure on the MoNuSeg and TNBC datasets and show quantitatively and qualitatively that our proposed method improves object-level metrics substantially. 
% Code for this paper will be made available after publication.
We consider the problem of segmenting cell nuclei instances from Hematoxylin and Eosin (H\&E) stains with weak supervision. 
While most recent works focus on improving the segmentation quality, this is usually insufficient for instance segmentation of cell instances clumped together or with a small size. 
In this work, we propose a two-step post-processing procedure, \emph{Split} and \emph{Expand}, that directly improves the conversion of segmentation maps to instances.
In the \emph{Split} step, we split clumps of cells from the segmentation map into individual cell instances with the guidance of cell-center predictions through Gaussian Mixture Model clustering.
In the \emph{Expand} step, we find missing small cells using the cell-center predictions (which tend to capture small cells more consistently as they are trained using reliable point annotations), and utilize Layer-wise Relevance Propagation (LRP) explanation results to expand those cell-center predictions into cell instances.
Our \emph{Split} and \emph{Expand} post-processing procedure is \textit{training-free} and is executed at inference-time only.
To further improve the performance of our method, a feature re-weighting loss based on LRP is proposed. 
We test our procedure on the MoNuSeg and TNBC datasets and show that our proposed method provides statistically significant improvements on object-level metrics.
Our code will be made available.

\end{abstract}

\section{Introduction}
% \lgg{Maybe motivate the problem a little more?}

% \lgg{emphasize on the problem of semantic -> instance segmentation. and how the boundaries might be missed. and how it misses small cells through the processing of the output map}

% \lgg{emphasize more on the intuition behind the explainability technique -> the explanation for the existence of cell centers, should be the entire cell itself.}

% \lgg{possibly can add a diagram to explain the leap from semantic segmentation to instance segmentation.}

Instance segmentation is crucial in many biomedical applications, such as phenotyping \cite{phenotype}, cell-tracking \cite{cell_tracking} and computer-aided cancer diagnosis \cite{breast_cancer_diagnosis,Beck2011SystematicAO}.
% Deep learning models can potentially be used to master this task \cite{DLtechniques}, leading to improvements in those applications, but require large amounts of high quality annotated data.
Deep learning models can potentially be deployed to improve instance segmentation quality \cite{DLtechniques} and increase the reliability of these applications, but require large amounts of high quality annotated data.
% Moreover, getting fully annotated datasets of medical images is \textit{expensive} \cite{qu2019_weakly,bearman2016s,Qu_2020}. 
Moreover, getting fully annotated datasets of medical images is \textit{expensive} \cite{qu2019_weakly,bearman2016s,Qu_2020},
% as many pixels need to be carefully annotated for each individual cell.
e.g. for H\&E stains, many pixels need to be carefully annotated for each individual cell.
% , especially when many pixels need to be carefully annotated for each individual cell.
% many individual pixels need to be carefully annotated.
% each individual pixel needs to be annotated carefully.
% all pixels within all cells need to be annotated. 
It is thus costly for qualified medical experts to annotate large databases, yet alternatives such as cloud annotation are less reliable. 
% face problems w.r.t. their reliability.
% Using cloud annotation faces acceptance problems w.r.t. its reliability while it is time-consuming for qualified medical experts to annotate large databases. 
% Conversely, it is much easier to obtain \textit{dot-annotated} datasets, which is the primary motivating factor behind weakly supervised segmentation.
% Conversely, it is much easier to obtain \textit{dot-annotated} datasets as only one pixel in each cell needs to be annotated, which is the primary motivating factor behind weakly supervised segmentation.
Conversely, it is much easier to obtain \textit{dot-annotated} datasets (e.g. only one pixel in each cell needs to be annotated), which is the primary motivating factor behind weakly supervised segmentation.

The prevailing approach for weakly supervised (dot-annotated) cell segmentation problem consists of generating coarse labels using various methods (e.g. Voronoi partitioning and clustering), and training the segmentation model using these labels \cite{qu2019_weakly,guo2021learning,chamanzar2020weakly,tian_selfstimulated,yen2020ninepins}. 
While this approach works rather well for segmenting foreground pixels from the background, the conversion from the output segmentation map to the instance segmentation one is still coarsely done using basic morphological operations. 
% \lgg{In most cases, the boundaries between cells (which might require segmentation map confidence scores to determine) are not well handled,} 
In many cases, boundaries between cells are not well handled by these morphological operations,
leading to clumps of cells being mistakenly identified as a single instance. 
% Additionally, we also observe that small cells are often missed. 
% \lgg{Additionally, we also observe that small cells are often missed or removed by these morphological operations.}
Additionally, we also observe that small cells are often missed in the output segmentation map, which is likely due to the errors in the coarse labels (as noted in \cite{qu2019_weakly,guo2021learning,chamanzar2020weakly}), that might disproportionately affect the smaller cells more.
% that disproportionately affects the smaller cells more.
% This is potentially a big issue, especially when the number of cells detected is a primary concern, e.g cell counting.
Both of these are potentially critical issues, especially when the number of cells detected is a primary concern, e.g cell counting.
% \lgg{any particular reason why small cells are missed?}

% \lgg{explain motivation along with the method}
In our work, we tackle the aforementioned problems with a two-step post-processing procedure at inference time.
To tackle the \textit{mishandling of boundaries leading to clumps of cells identified as instances}, we introduce a \emph{Split} step that performs splitting of clumps of cells when multiple cell centers have been identified within a clump.
% We split the clumps into instances through clustering using a Gaussian Mixture Model, 
We split the clumps into instances through a Gaussian Mixture Model clustering, 
while taking into account the confidence of cell center prediction maps, such that pixels with lower prediction values within the clumps are likelier to be the boundaries between cells.
% We take into account the confidence of segmentation maps, such that pixels with lower confidence within the clumps are likelier to be the boundaries between cells.
% To tackle the \emph{problem of missing small cells}, we introduce an \emph{Expand} step that performs the expansion of predicted cell centers (that are not identified in the segmentation map) into entire cell instances.
To tackle the \emph{problem of missing small cells}, we introduce an \emph{Expand} step where our model predicts the locations of cell centers, and performs the ``expansion" of some predicted cell centers (those not identified in the segmentation map) into entire cell instances.
% Here, we adopt an explanation method, LRP, to identify entire small cells using predicted cell centers. 
Crucially, the supervision for the cell-center (CC) prediction task comes from the \textit{reliable dot annotations} (and not the processed coarse labels), which makes these CC predictions more likely to be accurate than the segmentation map.
% and is likely to be more accurate.
We adopt an explanation method, LRP, to identify entire small cells from predicted cell centers. 
% The intuition is that the explanation for the cell center predictions corresponds to the identification of a cell instance (and LRP will produce a heatmap that highlights that cell).
% Intuitively, each \textit{cell center prediction} from our model should be triggered by the identification of a \textit{cell instance around that point}
% Intuitively, \textit{each point predicted by our model to be a cell center} should have been triggered by the identification of a \textit{cell instance around that point} (the heatmap of which is produced by LRP).
% Intuitively, \textit{each point predicted by our model to be a cell center} should have been triggered by the identification of a \textit{cell instance around that point}, and \textit{LRP produces a heatmap that precisely highlights that cell instance} to explain the cell center prediction.
Intuitively, each point predicted by our model to be a cell center is predicted as such due to an identified \textit{cell instance around that point}, and \textit{LRP produces a heatmap that precisely highlights that cell instance} to explain the cell center prediction.
% (and thus, explained) by the identification of a cell instance (and LRP will produce a heatmap that highlights that cell).
%triggered, caused
% will produce a heatmap that corresponds to the entire (small) cell.
% the explanation for the pixels (classified as cell centers) will produce a heatmap that corresponds to the entire (small) cell.
LRP is capable of identifying the inputs that are related to the predictions and provides high-quality explanations and visualizations in many evaluation studies \cite{COMPARISON:poerner2018evaluating,evaluatingthevisualization,EVALLSTMACL:arrasACL2019,ASSESS:Lapuschkin2019}. 
To accomplish the post-processing procedure above, we require the prediction of CCs, thus we add another head to our segmentation model that does CC prediction, and simultaneously train both heads (segmentation and CC-prediction) during training.
% However, the CC-prediction task is challenging as the positive labels are sparse compared to the background. 
However, the CC-prediction task is challenging as CCs are sparse compared to the background. 
To get better accuracy on the CC output head, we propose a feature re-weighting (FRW) loss based on explanation methods inspired by \cite{sun2021explanation}.
% This loss re-weights features using explanation scores, such that features that contribute more towards the rare class (positive CC prediction) are up-weighted and trained more, leading to better training on the imbalanced dataset.
This loss re-weights features using explanation scores, such that features that contribute more towards the rare class (positive CC prediction) are up-weighted, leading to better training on the imbalanced dataset.

To summarize, our contributions are:
    
    \textbf{1}. We propose a novel post-processing procedure, \emph{Split and Expand}, for cell instance segmentation.
    \emph{Split and Expand} is training-free and model-agnostic, and improves cell instance segmentation by resolving the clumps of cells and missing small cells in the segmentation map. 
    To our best knowledge, we are the first to segment cell instances with the guidance of explanation results. 
    
    % \textbf{2}. We employ cell center predictions to split segmented objects with multiple instances.
    
    % \textbf{3}. We explore using the explanations of cell center predictions to expand small cell instances. To our best knowledge, we are the first to segment cell instances with the guidance of explanation results. 
    
    % \textbf{2}. To overcome the label imbalance when training the cell-center output head, we propose a feature re-weighting loss that leads to further improved performance.
    \textbf{2}. To overcome the label imbalance when training the cell-center output head, we propose a feature re-weighting loss that leads to further improvements.    
    % which leads to further improved object-level performance. 
    
    % \textbf{3}. Experiments on MoNuSeg \cite{monuseg} and TNBC \cite{naylor_TNBC} datasets demonstrate the effectiveness of our proposed methods on object-level metrics over the baseline. 

    % \textbf{3}. Experiments on MoNuSeg \cite{monuseg} and TNBC \cite{naylor_TNBC} datasets demonstrate the effectiveness of our proposed methods  over the baseline. 

    \textbf{3}. Experiments on MoNuSeg \cite{monuseg} and TNBC \cite{naylor_TNBC} datasets show that our proposed methods provide statistically significant improvements over the baseline. 

% We test all our procedures on the MoNuSeg \cite{monuseg} and TNBC \cite{naylor_TNBC} datasets.

% \begin{itemize}
%     \item \emph{Split}: At inference time, adding a simple yet novel Instance-Splitting procedure using the predicted CC's substantially improves object-level metrics. 
%     \item \emph{Expand}: We propose CC-Expansion, where at prediction time we "expand" the CC's using LRP explanations that we qualitatively observe to work well on detecting the smaller cells. To the best of our knowledge, this is novel in the field of cell instance segmentation.
%     \item We show that applying a FRW loss on training of the CC output head leads to better object-level metrics on the instance segmentation task.
% \end{itemize}

\section{Related Work}
\textbf{Weakly supervised cell segmentation} is the task where a cell segmentation model is trained using weak supervision (usually dot-annotated data), and has been receiving more attention recently due to its practical utility.
% is an important task that has been receiving more attention recently.
% Several methods have been proposed to tackle the weakly-supervised cell segmentation problem. 
% Notably, \cite{qu2019_weakly} pre-processed point labels into Voronoi cells and clusters based on color and used them as coarse labels. \cite{yoo2019pseudoedgenet} proposed an edge network that could learn an edge map. \cite{chamanzar2020weakly} proposed a repel encoding loss that was applied in an alternating manner to the other losses. \cite{Obikane_2020} posed a weakly-supervised domain adaptation problem and proposed an adversarial training method to tackle it. More recently, \cite{Qu_2020} proposed a weakly-supervised partial-annotations paradigm and self-supervised methods that perform well on that problem. \cite{tian_selfstimulated} also proposed a self-supervised learning method together with a contour refinement technique. 
% \cite{guo2021learning} proposed mask-guided attention and label refinement.
% \cite{yen2020ninepins} proposed to additionally use coarse distance labels.
Notably, \cite{qu2019_weakly} pre-processed point labels into Voronoi cells and clusters based on color and used them as coarse labels,
\cite{yoo2019pseudoedgenet} proposed to train an edge network,
\cite{chamanzar2020weakly} proposed a repel encoding loss,
\cite{yen2020ninepins} proposed to use coarse distance labels and
\cite{Obikane_2020} proposed an adversarial training method. 
% \cite{chamanzar2020weakly,Obikane_2020} further proposed new training methods, while \cite{Qu_2020,tian_selfstimulated,guo2021learning} proposed some self-supervised methods
Recently, \cite{tian_selfstimulated,guo2021learning} further proposed new methods involving self-training. 
% More recently, \cite{Qu_2020} proposed a weakly-supervised partial-annotations paradigm and self-supervised methods that perform well on that problem. \cite{tian_selfstimulated} also proposed a self-supervised learning method together with a contour refinement technique. \cite{guo2021learning} proposed mask-guided attention and label refinement.
Throughout these works, the instance segmentation output is obtained by applying morphological operations on the output segmentation map. 
% \lgg{
% Differently, our Split and Expand method not only uses morphological operations, but also refines the instances by \textit{breaking up clumps} into smaller instances and \textit{get new instances} using explanation results from LRP.
% To the best of our knowledge, we are the first work to utilize explanation methods to guide cell instance segmentation, whether it be our Expand step, or the FRW loss.
% We note that our work is complementary to the others proposed above (which generally focus on the training stage), and can be easily added onto their methods.}
Differently, our Split and Expand method further refines cell instances by \textit{splitting up clumps} into smaller instances, \textit{finding small instances} through CC-prediction (trained with reliable dot annotations) and \textit{expanding them} using explanation results from LRP.
% To the best of our knowledge, we are the first to tackle these issues with contemporary cell instance segmentation methods.
To the best of our knowledge, our method is the first to tackle these issues.
% with contemporary cell instance segmentation methods.
%
% To the best of our knowledge, we are the first work to utilize explanation methods to guide cell instance segmentation, whether it be our Expand step, or the FRW loss.
We further note that our work is complementary to the others proposed above (which generally focus on the training stage), and can be \textit{easily added onto their methods}.
% by adding a CC output head.
% \lgg{shorten existing ones?}

% In \cite{graham2019hover,chen2019instance,yen2020ninepins}, a distance-map approach was applied to separate instances. Among them, only \cite{yen2020ninepins} considered the dot-annotated case, and is most closely related to our work. However, their approach is different from ours as they generate a new coarse label that approximates the distance labels and also requires training on those labels using the same architecture as \cite{graham2019hover}. 

% The post-processing part of our work also bears some similarity to seeding methods mostly used in semantic segmentation methods with global class labels \cite{kolesnikov2016seed,ahn2019weakly,zhang2020causal}. These methods generally use class activation mappings (CAMs) \cite{zhou2015cam} to localize foreground objects in order to perform self-supervised iterative learning. Different from the above methods, our Split and Expand is training-free, where we employ the CC prediction results (splitting) and explanation results of LRP (expansion) during test time.

% Jiamei: I will add one para of this
\textbf{LRP} is one of the explanation methods that aim at de-mystifying the black box of deep neural networks (DNN) and interpreting the model decisions \cite{LRP:bach2015pixel}. 
% The explanations results reflect the contribution of a neuron to the model decision, which is helpful for understanding and debugging the models\cite{ASSESS:Lapuschkin2019,AnaImg:anders2019analyzing}. 
The explanation results reflect the contribution of a neuron to the model decision, which is helpful for understanding the models \cite{ASSESS:Lapuschkin2019,AnaImg:anders2019analyzing}. 
Recently, several works studied the applications of explanation results \cite{Grad-Camloss:halliwell2020trustworthy,sun2021explanation} in other domains. 
% In this work, we explore using LRP explanations to both serve as a cell segmentation map, and also to design a FRW loss for better CC-prediction.
To our best knowledge, we are the first work to utilize explanation methods to guide cell instance segmentation, whether it be our Expand step, or the FRW loss.

\section{Method}
\subsection{Data pre-processing}
% Using the point labels, we generate three different types of coarse labels: Cluster, Voronoi, Enlarged Point. The Cluster labels $GT_{C}$ and Voronoi labels $GT_{V}$ are processed similar to \cite{qu2019_weakly}. Briefly speaking, the Cluster labels are a result of using K-Means on pixel values to get three clusters of pixels, which can then be assigned to be Cell, Background or Ambiguous classes. The Voronoi labels are obtained by applying Voronoi partitioning on the point labels. Lastly, the Enlarged Point labels $GT_{P}$ are obtained by expanding each point label into a 3x3 square. This helps provide stronger supervision for the model.
% Other alternatives are possible, including expanding to a larger square, or applying a Gaussian function as in \cite{Obikane_2020,Qu_2020}, but we find that simple 3x3 square labels suffice in training the CC outputs.
% \lgg{somehow abstract the other labels away, as coarse labels?? dont focus on only cluster and voronoi labels.... because other methods might have other coarse labels. And also because they might ask about our usage of cluster labels.}

We pre-process our dot-annotated labels into various other forms to train both the segmentation head and the CC-prediction head. 
% In order to train the CC-prediction head, we require Enlarged Point labels $GT_{P}$ which are obtained by expanding each point label into a 3x3 square. This helps provide stronger supervision for the model.
%

To train the CC-prediction head, we require Enlarged Point labels $GT_{P}$ which are obtained by expanding each point label into a 3x3 square. 
$GT_{P}$ is reliable, and helps provide stronger supervision for the model than point labels.

% Additionally, we also use a FRW loss $L_{FRW}$ with respect to the Enlarged Point labels. The weights of the features are generated with the explanation results of LRP.

% \begin{figure}[htb]
\begin{figure}
\centering
\includegraphics[width=0.8\textwidth]{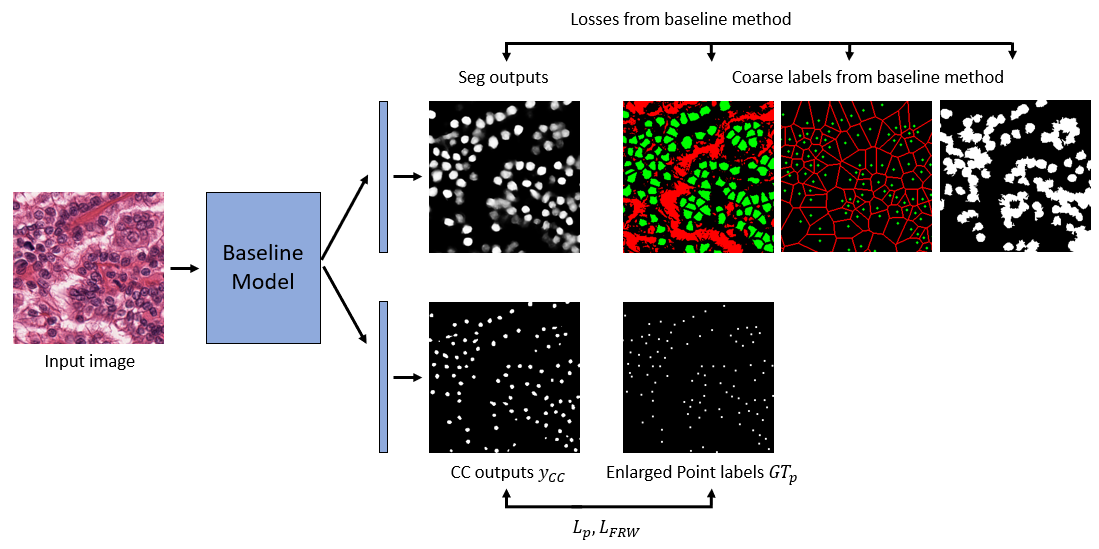}
\vspace{-0.6cm}
\caption{An overview of our training method, network outputs, labels and losses.
% For visualization purposes, we show the coarse labels (Cluster and Voronoi) as used in \cite{qu2019_weakly}.
For visualization purposes, we show the Cluster, Voronoi and Superpixel coarse labels. We emphasize that our method can be\textit{ easily applied onto most existing segmentation baseline methods}, by simply adding a CC output head to the baseline model.
% \lgg{with \cite{qu2019_weakly} as the baseline method.} \lgg{Note that our method is model-agnostic and the coarse labels do not need to be cluster and Voronoi labels}
\label{training_fig}}
\vspace{-0.6cm}
\end{figure}

To train the segmentation head, we generate the coarse labels that our baseline model needs, e.g. when using \cite{qu2019_weakly} as the baseline, we need Cluster and Voronoi labels.
% For example, when using \cite{qu2019_weakly} as the baseline, we need Cluster and Voronoi labels.
% We highlight that our post-processing method is model-agnostic, and can work effectively even when different types of coarse labels are used here.
We highlight that our post-processing method is model-agnostic, and can work effectively on different baselines and coarse labels.
% of models and coarse labels.

% For the segmentation model, we adopt whatever coarse labels a baseline model uses and preprocess whatever data they require. Using the point labels, we generate three different types of coarse labels
% For example, when using \cite{qu2019_weakly} as the baseline, we need cluster and Voronoi labels.

\subsection{Cell-center predictions and feature re-weighting loss}
A summary of our proposed network and losses can be seen in Figure \ref{training_fig}. As our method focuses on post-processing, we can adopt any baseline model, loss, or labels for the segmentation output. 
Moreover, we adapt the model to also generate an extra CC output $y_{CC}$ from its features.
The CC output is trained using pixel-wise cross entropy loss with respect to the Enlarged Point labels $GT_P$. 
Crucially, $GT_P$ is reliable as it is is annotated by experts, and thus the CC prediction tends to perform better than segmentation (especially on smaller cells where the coarse labels might contain more errors).
Moreover, we note that there is a large label imbalance of CC pixels compared to background pixels in $GT_P$, that might affect performance.
To further improve the training of the CC output $y_{CC}$ under this label imbalance, we additionally propose a FRW loss $L_{FRW}$ with respect to the Enlarged Point labels $GT_P$, which we describe next.

A core part of our FRW loss is LRP, which is capable of explaining the decisions of various DNNs \cite{LRP:bach2015pixel,LRP-LSTM:arras2017explaining,GNNLRP:schnake2020xai,Explainkmeans:kauffmann2019clustering}. It assigns an explanation score to every neuron that reflects supporting (positive scores) or opposing (negative scores) contribution to the predictions \cite{ASSESS:Lapuschkin2019}. Furthermore, compared to other gradient-based explanation methods, LRP explanation scores reflect more of the related features that are used by the model to make decisions, which has been evaluated in \cite{COMPARISON:poerner2018evaluating,evaluatingthevisualization,EVALLSTMACL:arrasACL2019}. Thus, we apply the LRP explanations to design the feature re-weighting loss, $L_{FRW}$. In other domains, such explanation-guided losses have been known to perform well on small sample sizes or imbalanced datasets in other domains \cite{sun2021explanation,ImageCAP:sun2020understanding}, and the latter presents itself in our CC prediction task.

Let $\bm{f}_l$ denote the feature map of layer $l$ in our model. We first perform a forward pass through the model to generate an original prediction $y_{CC}$. We then explain the generated prediction with LRP and obtain the explanation scores of the feature map $R(\bm{f}_l)$. We refer to the $LRP_{\alpha1}$ rule to calculate $R(\bm{f}_l)$, as suggested in \cite{SebasIJCNN2020:kohlbrenner2019towards}. The re-weighted feature $\hat{\bm{f}}_l$ is calculated as follows:
\begin{equation}
    w(\bm{f}_l) = l_{norm}(R(\bm{f}_l)) + 1, \hat{\bm{f}}_l = w(\bm{f}_l) \odot \bm{f}_l  \label{equ: re-weightfeature}
\end{equation}
% where $l_{norm}$ is a normalization layer where we divide using the maximum absolute value, $w(\bm{f}_l)$ is the generated weight for feature map $\bm{f}_l$, $\odot$ is the Hadamard product operation, and $\hat{\bm{f}}_l$ is the re-weighted feature. The re-weighted feature $\hat{\bm{f}}_l$ is then fed forward to obtain a new output $\hat{y}_{CC}$, and the cross entropy loss is calculated with respect to the Enlarged Point labels. 
where $l_{norm}$ is a normalization layer where we divide using the maximum absolute value, $w(\bm{f}_l)$ is the generated weight for feature map $\bm{f}_l$, $\odot$ is the Hadamard product operation, and $\hat{\bm{f}}_l$ is the re-weighted feature. The re-weighted feature $\hat{\bm{f}}_l$ is then fed forward to obtain a new output $\hat{y}_{CC}$, and the cross entropy loss is calculated with respect to the Enlarged Point labels.

% \todo{Introduce how to calculate $R(\bm{f}_l)$ with lrp rules or cite other papers like \cite{FewShotLRP:sun2020explanation,SebasIJCNN2020:kohlbrenner2019towards}}

% \todo{A shot para to explain the underling idea of the re-weighting, that is, the re-weighting upscale the more related features and those related features receive higher loss during training.}

With the re-weighting operation, we have $w(\bm{f}_l)>1$ for the features with positive LRP explanation scores (indicating support) and $w(\bm{f}_l)<1$ for those with negative LRP explanation scores (indicating opposition) \cite{ASSESS:Lapuschkin2019,sun2021explanation}. 
% Thus, the re-weighted features up-scale the related parts that are tuned more with $L_{FRW}$.     
Thus, the re-weighted features up-scale the related parts and are tuned more with $L_{FRW}$.

% Our final loss looks like this:
% \begin{equation}
%     Loss = \alpha_{Voronoi} L_{Voronoi} + \alpha_{Cluster} L_{Cluster} + \alpha_{Point} L_{Point} + \alpha_{FRW} L_{FRW}
% \end{equation}
% Our final losses for each output head is as follows, where $\alpha_{V},\alpha_{C},\alpha_{P},\alpha_{FRW}$ represent the weights of the Voronoi, Cluster, Point and FRW losses:
% \begin{align}
%     Loss_{seg} &= L_{V} + L_{C} = \alpha_{V} CE(y_{seg}, GT_{V}) + \alpha_{C} CE(y_{seg}, GT_{C}) \\
%     Loss_{CC} &=  L_{P} + L_{FRW} = \alpha_{P} CE(y_{CC}, GT_{P}) + \alpha_{FRW} CE(\hat{y}_{CC}, GT_{P})
% \end{align}

Our final loss for the CC output head is as follows, where $\alpha_{P},\alpha_{FRW}$ represent the weights of the Point and FRW losses:
\begin{align}
    % Loss_{seg} &= L_{V} + L_{C} = \alpha_{V} CE(y_{seg}, GT_{V}) + \alpha_{C} CE(y_{seg}, GT_{C}) \\
    Loss_{CC} &=  L_{P} + L_{FRW} = \alpha_{P} CE(y_{CC}, GT_{P}) + \alpha_{FRW} CE(\hat{y}_{CC}, GT_{P})
\end{align}

\subsection{Post-processing: Split and Expand}
\begin{figure}
\includegraphics[width=\textwidth]{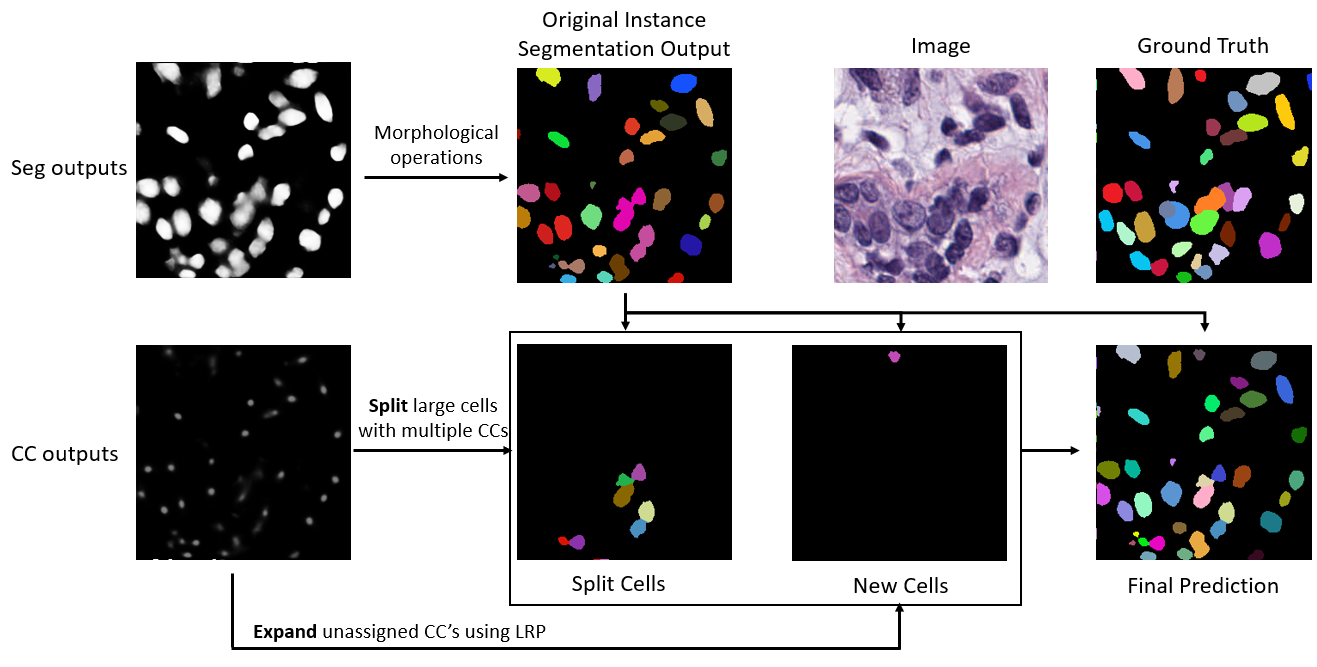}
% \caption{An overview of our two-step procedure of Instance-Splitting and CC-Expansion.}
\vspace{-0.8cm}
\caption{An overview of our two-step procedure of Split and Expand.
The Split step splits clumps of cells in the original instance segmentation output into smaller instances (e.g. the pink clump at the center is split into separate brown, green and purple instances).
The Expand step ``expands" unassigned CC's using LRP into cell instances (e.g. a ``new" small pink cell is shown here).
% These refinements are merged with the original instance segmentation output to form a final prediction.
% After obtaining the original instance segmentation and CC outputs from our model, we first 
We highlight that this procedure is \textit{training-free}.
}
\label{testing_fig}
\vspace{-0.6cm}
\end{figure}

At test time, we obtain our Segmentation output and perform morphological operations (remove small objects, fill holes) to get an instance segmentation map. We then use our CC output to conduct our two-step post-processing (Split and Expand) on the instance segmentation map.
% To the best of our knowledge, this post-processing method is novel in the field of cell instance segmentation. 
% \lgg{The Split step helps to split clumps of cells into instances, while the Expand step helps to identify small cells, which are often missed in the sgementation map.}
% The Split step helps to split clumps of cells into instances, while the Expand step helps to expand the small cells (whose CCs have been predicted), which are often missed in the segmentation map.
The Split step helps to split clumps of cells into instances, while the Expand step helps to materialize the small cells (whose CCs have been predicted), which are often missed in the segmentation map.
% We emphasize that we do not train on the post-processing task explicitly.
We emphasize that we \textit{do not train on the post-processing task explicitly}.
An overview of this process can be seen in Figure \ref{testing_fig}.

% \lgg{explain more. maybe can make into subsections. add some equations? especially for the splitting. add more motivations for each step. overlapping cells and small cells.}
% The first step is the Instance-Splitting step. 
% The first step is the \emph{Split} step. 
The first step is the \emph{Split} step, which is a short form for Instance-Splitting. 
We first condense all blobs of CC outputs into single CC points by applying the same morphological operations and taking the center of each blob instance. Then, we find instances in our instance segmentation map that cover two or more CC points. 
% Intuitively, these identified instances are \textit{clumps} that contain two or more cells in close proximity. 
Intuitively, these identified instances are \textit{clumps} that contain two or more cells in close proximity, e.g. a clump containing $k$ CCs is likely to contain $k$ cell instances. 
% \reh{Next, we perform the split step by weighted-sample clustering based on CC predictions.
% By observation, we find that the CC probability map is normal about each cell center, forming ellipsoids on the feature map. Hence, we model the set of CC probabilities near overlapping regions using a multivariate gaussian mixture model. After using expectation maximization to refine relevant parameters, we use the model to classify parts of the cell into its clusters. We find this method to work well over hard clustering methods or splitting based on distance metrics.}
% Next, we perform weighted-sample clustering based on CC predictions within these clumps, to split them into $k$ smaller instances. 
Next, we perform weighted clustering based on CC predictions within these clumps to split them into $k$ smaller instances, where \textit{each cell instance is a cluster of pixels}. 
% As most cells have a roughly ellipsoidal shape, we choose to cluster using a bi-variate Gaussian Mixture Model, that is capable of modelling (2D) ellipsoids by fitting appropriate covariance values.
As most cells have a roughly ellipsoidal shape, we adopt a bi-variate Gaussian Mixture Model, that is capable of modelling (2D) ellipsoids by fitting appropriate covariance values.
More specifically, the algorithm will cluster all points within the clump into $k$ (ellipsoidal) clusters, where each point is weighted by the (non-negative) CC prediction values.
As such, pixels with lower CC prediction values are likelier to be points on the cell's boundary, while pixels with higher CC prediction values are likelier to be at the center of the cell.
Overall, this step can split clumps successfully into instances that match the ground truth well, as shown in Figure \ref{testing_fig}.
%  Figure \ref{testing_fig}, clumps are successfully split into instances that match the ground truth well.
We find this method to work well over other clustering methods or splitting based on distance metrics.
% \lgg{find somewhere to refer to the figure?}
% We find this method to work well over hard clustering methods or splitting based on distance metrics.
%
%
%
% We split the clumps into instances through a Gaussian Mixture Model clustering, 
% while taking into account the confidence of cell center prediction maps, such that pixels with lower confidence within the clumps are likelier to be the boundaries between cells.
%
%
%
% We perform a simple procedure to split these clumps up, by assigning each point in the clump to its nearest CC point in L1 distance. While other distance measures and procedures can be considered, in our experiments the simple L1 distance seems to work rather well. Note that this step does not change any pixel-level metrics as it just splits existing instances. This step works well to improve object metrics when cells are more likely to be in clumps.

% In the CC-Expansion step, we expand CC predictions that are not contained in any instances in our original segmentation map. 
% In the \emph{Expand} step, we expand CC predictions that are not contained in any instances in our original segmentation map. 
% In the \emph{Expand} step, we first identify CC predictions that are not contained in any instances in our original segmentation map. 
% Intuitively, these represent CC points of cells that should have been identified but were missed in the segmentation map.
Next is the \emph{Expand} step, which is short for CC-Expansion.
We first identify CC predictions that are not contained in any instances in our original segmentation map --
intuitively, these represent CC points of cells that should have been identified but were missed in the segmentation map.
We next ``expand" these CC points into entire cell instances.
Using a single backward pass in LRP on the identified blobs in the CC outputs, we obtain an "explanation" heatmap of the positive CC predictions, which closely resembles the shape of those cells. 
Intuitively, this is because each positive CC prediction should be explained by the presence of a cell instance around it (which is materialized using LRP).
% After applying a heatmap threshold, an approximate instance segmentation of the cells in the original image is obtained. 
After applying a heatmap threshold, we obtain the instance segmentation map of these cells.
We note that this method is \textit{efficient} as \textit{only one LRP backward pass is needed per input sample}.
Furthermore, in order to prevent repeat cells or generating new clumps of cells, an overlap threshold ($\frac{|Cell_1 \cap Cell_2|}{min(|Cell_1|,|Cell_2|)}$) is implemented to discard instances with large overlaps with any existing cells. 
% This situation happens occasionally on some oddly-shaped cells or clumps of cells.

A Split \& Expand algorithm summary can be found in the Supplementary.

% \begin{equation}
%     Overlap = \frac{|Cell_1 \cap Cell_2|}{min(|Cell_1|,|Cell_2|)}
% \end{equation}

% As the difference in cell sizes can be very large, we find that the overlap threshold ($\frac{|Cell_1 \cap Cell_2|}{min(|Cell_1|,|Cell_2|)}$) works better than the IoU threshold which is more commonly used. In practice, we find that the cells found by this step are often small, obscure and sometimes even unlabelled in the ground truth annotations. We discuss this further in the results section. \lgg{possibly supplementary}

\section{Experiment details}
% For the baseline model, we used the same U-Net architecture as \cite{buda_brainseg}, and used their pre-trained weights throughout our experiments. Notably, this U-Net does not use a ResNet encoder unlike \cite{qu2019_weakly,tian_selfstimulated}, and we get a similar level of accuracy. For the CC output branch, two additional 3x3 Conv-BatchNorm-Relu layers are added for better performance. 
% We experiment using two baseline methods  \cite{qu2019_weakly,guo2021learning} that use different models, losses and coarse labels for training the segmentation head.
We experiment using two baseline methods  \cite{qu2019_weakly,guo2021learning} that use different models, losses and coarse labels (for training the segmentation head).
% \lgg{Did we use pretrained weights from \cite{buda_brainseg}? Thats what i did, and the architecture is different from the original paper.... And, did we use pretrained weights from \cite{guo2021learning}?} 
% \reh{Yes, used pretrained weights from \cite{buda_brainseg}, no did not use pretrained weights from \cite{guo2021learning}}
% , and we follow their methods entirely for the segmentation head.
% In particular, we add the CC output branch before the second last layer of both baseline models (i.e. the U-Net for \cite{qu2019_weakly} and two Feature Aggregation Networks for \cite{guo2021learning}), which mirrors the last two layers of the segmentation branch.
We duplicate the last two layers of both baseline models to form the CC output branch.
% We add the CC output branch before the second last layer of both baseline models (i.e. the U-Net for \cite{qu2019_weakly} and two Feature Aggregation Networks for \cite{guo2021learning}), which mirrors the last two layers of the segmentation branch.
% \reh{We used same model architectures as our baselines, U-Net \cite{qu2019_weakly} and two Feature Aggregation Networks detailed in \cite{guo2021learning}.}
% We used the U-Net architecture and pre-trained weights from \cite{buda_brainseg} throughout our experiments. 
% Notably, this U-Net does not use a ResNet encoder unlike \cite{qu2019_weakly,tian_selfstimulated}, and we get a similar level of accuracy. 
% \reh{The CC output branch is added before the second last layer, mirroring the Conv-BatchNorm-Relu layers of the segmentation branch.}
% For the CC output branch, two additional 3x3 Conv-BatchNorm-Relu layers are added for better performance. 

Each model was trained for 150 epochs using an Adam optimizer with a learning rate of 0.001. 
% For all configurations, 
% $\alpha_{V} = 0.25, \alpha_{C} = 0.25$. 
We set $\alpha_{P}=0.5, \alpha_{FRW}=0.5$ when $L_{FRW}$ is used, otherwise $\alpha_{P}= 1.0$. 
% \reh{For experiments using FRW loss, we applied it at the first encoding layer.}
The FRW loss is computed using the feature map at the first encoding layer.
We set our small object threshold at 10, CC confidence threshold at 0.05, heatmap threshold at 0.1 and overlap threshold at 0.5. 
Images were augmented using random resizing, rotations, flips, crops and affine transformations.
% Images were cropped into 224$\times$224 images during training, and then augmented using random resizing, rotations, flips, crops and affine transformations.

% All other training and testing settings follow

% At the testing phase, we set our CC confidence threshold to be 0.1, heatmap threshold to be 0.05 and overlap threshold to be 0.5. 
% We found these values to work in general and they did not require much tuning. In principle, we can tune these values using our best model on the validation set before testing, but we would like to keep our procedure simple. 

% \lgg{follow the evaluation method}
% Following the experimental procedure of \cite{guo2021learning,tian_selfstimulated,yoo2019pseudoedgenet},
% we evaluated our methods on the MoNuSeg (Multi-Organ Nucleus Segmentation) dataset \cite{monuseg} and the TNBC (Triple Negative Breast Cancer) dataset \cite{naylor_TNBC}. These datasets are publicly available. MoNuSeg contains 30 fully-annotated 1000x1000 H\&E stained histology images of different organs. TNBC contains 50 fully-annotated 512x512 H\&E stained histology images from different parts of tissue of patients with the same cancer type. Following the procedure of \cite{guo2021learning,tian_selfstimulated,yoo2019pseudoedgenet}, we perform 10-fold cross-validation on both of them, taking the ratio of training:validation:test to be 8:1:1. As computing the object-level metrics is computationally intensive, during cross-validation we compute pixel-level metrics once every 10 epochs and select the model with the best pixel-level F1 score.
Following the experimental set-up of \cite{guo2021learning,tian_selfstimulated,yoo2019pseudoedgenet},
we evaluated our methods on two publicly available datasets: the MoNuSeg (Multi-Organ Nucleus Segmentation) dataset \cite{monuseg} and the TNBC (Triple Negative Breast Cancer) dataset \cite{naylor_TNBC}. MoNuSeg contains 30 fully-annotated 1000x1000 H\&E stained histology images of different organs. TNBC contains 50 fully-annotated 512x512 H\&E stained histology images from different parts of tissue of patients with the same cancer type. Following the procedure of \cite{guo2021learning,tian_selfstimulated,yoo2019pseudoedgenet}, we perform 10-fold cross-validation on both of them, taking the ratio of training:validation:test to be 8:1:1. 

% As computing the object-level metrics is computationally intensive, during cross-validation we compute pixel-level metrics once every 10 epochs and select the model with the best pixel-level F1 score.

% \lgg{rui en, please work on this part. importantly, add the new baselines from the newest paper. And also, we can remove comparison with ninepins (which is highlighted also)}
% \lgg{remember to define new small cells metrics}
\section{Results and Discussion}
% We report our experiment results on the MoNuSeg in Table \ref{monuseg_tab} and TNBC in Table \ref{tnbc_tab}. 
% \reh{We report 4 metrics: object-level DICE, AJI, small cell object-level DICE and small cell AJI.}
% We report 4 scores: Pixel-wise accuracy, Pixel-wise F1 score, Object-level DICE and AJI. 
In our experiments, we report 4 metrics: AJI, object-level DICE, small-cell AJI and small-cell DICE.
% Object-level DICE is a weighted sum of pixel-wise F1 scores between each ground truth cell and the best assigned segmented object, as well as between each segmented object and the best assigned ground truth, with the weights being the size of the ground truth/segmented object. 
\textbf{AJI} is a Jaccard overlap between the set of ground truths and segmented instances, calculated across the pairwise union and intersection between each ground truth and its assigned ``best" instance. 
\textbf{Object-level DICE} (DICE) is a weighted sum of pixel-wise F1 scores between each ground truth cell and the best assigned segmented instance, as well as between each segmented instance and the best assigned ground truth cell,\textit{ weighted by the size of the cell or instance}. 
Both AJI and DICE are important metrics adopted in previous works to measure instance segmentation quality \cite{qu2019_weakly,tian_selfstimulated,chamanzar2020weakly}, and can take values between 0 (worst) and 1 (best).
% \lgg{Moreover, these metrics place more importance on large cells, , and we found them generally insensitive to improvements in our \emph{Split} step.}
% \lgg{Moreover, these metrics are more affected by large cells compared to small cells, and small instances only minimally affect them, which we found to be inappropriate to quantify the improvements from our Expand step.}
% However, these metrics are more affected by large cells compared to small cells, and \textit{small instances only minimally affect them}, which we found to be inadequate to fairly quantify the improvements from our Expand step (where large cells are generally untouched).
However, these metrics are more affected by large cells compared to small cells, and \textit{small instances only minimally affect them}, which means that they are inadequate to fairly quantify the improvements from our Expand step (where large cells are generally untouched).
Thus, we report two new metrics:
% \textbf{small-cell DICE} and \textbf{small-cell AJI}, which are equivalent to Object-level DICE and AJI applied on the smallest 40\% of cells in each dataset. 
% Specifically, we calculate a pixel size threshold based on the $40^{th}$ percentile of cell size in each dataset, and only compute the metrics for ground truth cells or segmented instances with sizes below that threshold.
\textbf{small-cell AJI} and \textbf{small-cell DICE}, which are equivalent to AJI and DICE applied on small cells only. 
Specifically, we define a pixel size threshold (which we set to $300$), and only compute the metrics for ground truth cells or segmented instances with sizes below that threshold.
% find ``best" assignments for ground truth cells or segmented instances with sizes below that threshold.}
% segmented and ground truth instances larger than the threshold in pixels are removed before applying the same computations.
% \reh{Small cell DICE and AJI are equivalent metrics applied on cells smaller than the 20\% of cells in the datasets. Specifically, segmented and ground truth instances larger than the threshold in pixels are removed before applying the same computations.}
% Both AJI and Object-level DICE are object-level metrics that are important to making our case and can take values between 0 and 1. More information about Object-level DICE and AJI can be found in \cite{qu2019_weakly}.

%For each ground truth object, we assign to it a segmented object with the most overlap.

% The segmentation baseline is the baseline method taken from \cite{qu2019_weakly}.
We report our experiment results on the MoNuSeg in Table \ref{monuseg_tab} and TNBC in Table \ref{tnbc_tab}. 
The \textbf{Baseline} setting refers to our implementation of the segmentation baselines in \cite{qu2019_weakly} and \cite{guo2021learning},
% \reh{The segmentation baseline is the baseline method taken from \cite{qu2019_weakly} and \cite{guo2021learning}.}
the \textbf{Split} setting refers to the scenario where we only apply the \textit{Split} procedure on the segmentation baseline, and \textbf{Split \& Expand} refers to our full method.
% We evaluate the scenarios where we only apply the Instance-Splitting procedure (Split), and where we apply both the Instance-Splitting and CC-Expansion procedures (Split + Expand). 
We also compute the statistical significance (at 0.05 significance) of our improvements by performing a paired t-test between (Baseline and Split) and (Split and Split \& Expand)
% \reh{We perform a paired T-test between Baseline and Split, and between Split and Split \& Expand to check the significance of improvements at 5\% significance.}
% We also evaluate the effectiveness of the FRW loss by applying it on different layers of our U-Net (enc1, enc3, bottleneck). The layer that we use for the FRW loss is indicated in brackets.

\begin{table}
\caption{Split and Expand results on MoNuSeg (Top) and TNBC (Bottom) datasets. Best results are in bold. On all settings and metrics, our full Split and Expand method obtains significant improvements over the baseline method. Our Split step consistently provides significant improvements on the AJI and DICE metrics, while our Expand step consistently provides significant improvements on the small-cell metrics.
% \lgg{maybe can color the important boxes blue}
% \lgg{i think we should move the pixel-wise metrics into the supp material, for both datasets. or maybe dont even need to show.}
\vspace{-0.2cm}
}\label{monuseg_tab}
\centering

\begin{tabular}{|l|l|p{0.12\textwidth}|p{0.12\textwidth}|p{0.16\textwidth}|p{0.16\textwidth}|}
\hline
Methods & Configuration & AJI & DICE & AJI (small) & DICE (small) \\
\hline

\multirow{3}{*}{\begin{tabular}{@{}c@{}} Weakly\cite{qu2019_weakly} \end{tabular}} 
& Baseline & 0.488 & 0.692 & 0.206 & 0.384 \\
& Split & \cellcolor{blue!25}0.512* & \cellcolor{blue!25}0.708* & 0.209 & 0.391* \\
& Split \& Expand & \textbf{0.524**} & \textbf{0.710} & \cellcolor{green!25}\textbf{0.221**} & \cellcolor{green!25}\textbf{0.409**} \\
\hline

\multirow{3}{*}{\begin{tabular}{@{}c@{}} MaskGA\cite{guo2021learning} \end{tabular}} 
& Baseline & 0.482 & 0.705 & 0.196 & 0.379 \\
& Split & \cellcolor{blue!25}0.534* & \cellcolor{blue!25}0.728* & 0.214* & 0.402* \\
& Split \& Expand & \textbf{0.551**} & \textbf{0.732} & \cellcolor{green!25}\textbf{0.228**} & \cellcolor{green!25}\textbf{0.417**}\\
\hline
\end{tabular}
\end{table}

\begin{table}
\vspace{-0.3cm}
\label{tnbc_tab}
\centering

\begin{tabular}{|l|l|p{0.12\textwidth}|p{0.12\textwidth}|p{0.16\textwidth}|p{0.16\textwidth}|}
\hline

\multirow{3}{*}{\begin{tabular}{@{}c@{}} Weakly\cite{qu2019_weakly} \end{tabular}}
& Baseline & 0.516 & 0.698 & 0.199 & 0.372 \\
& Split & \cellcolor{blue!25}0.528* & \cellcolor{blue!25}0.706* & 0.203* & 0.377 \\
& Split \& Expand & \textbf{0.531} & \textbf{0.707} & \cellcolor{green!25}\textbf{0.215**} & \cellcolor{green!25}\textbf{0.387**} \\
\hline

\multirow{3}{*}{\begin{tabular}{@{}c@{}} MaskGA\cite{guo2021learning} \end{tabular}} 
& Baseline & 0.505 & 0.707 & 0.190 & 0.395 \\
& Split & \cellcolor{blue!25}0.532* & \cellcolor{blue!25}0.720* & 0.204* & 0.406* \\
& Split \& Expand & \textbf{0.533} & \textbf{0.721} & \cellcolor{green!25}\textbf{0.216**} & \cellcolor{green!25}\textbf{0.418**} \\
\hline
\end{tabular}

\begin{flushleft}
    % \small{* significant results from paired t-test between Baseline and Split \\ 
    % ** significant results from paired t-test between Split and Split \& Expand}
    \scriptsize{* significant results from paired t-test between Baseline and Split \\ 
    ** significant results from paired t-test between Split and Split \& Expand}
\end{flushleft}
\vspace{-2.5em}
\end{table}

Overall, our full \emph{Split} and \emph{Expand} method produces substantial gains \textbf{over all metrics and all settings}.
It is worthwhile to note that this is achieved without explicit training in our \emph{Split} and \emph{Expand} method.

Specifically, the Split step provides a significant improvement in object-level DICE and AJI in \textbf{all settings (highlighted in blue)}, demonstrating its effectiveness in splitting clumps of cell instances. 
We emphasize to the reader that this significant improvement comes from splitting clumps only, \textit{without changing the segmentation map}.
% and there is no change in the segmentation map.
% \reh{demonstrating the effectiveness of the bi-variate Gaussian Mixture Model in separating cell instances from clumps}.
Qualitative validation is provided in the Supplementary.
% This is supplemented by our qualitative observations of the images. An example can be seen in Figure \ref{testing_fig}, where several larger instances were split well. 
% A larger example can be seen in the supplementary material. 
% It is worthwhile to note that explicit training is not required in our \emph{Split} and \emph{Expand} method, to produce such substantial gains.

% \lgg{Although there is little basis for comparison, we carefully note that the DICE metric of our much-simpler method is slightly better than the fully weakly-supervised version in \cite{yen2020ninepins} on MoNuSeg. }

% \reh{Next, we observe that the CC-Expansion method improves small-cell DICE and AJI, demonstrating the effectiveness of LRP in identifying cells missed in the segmentation maps. Also, we find that the improvement in overall DICE and AJI is marginal in comparison, since most instances generated by LRP explanation are smaller cells.}
% Next, we observe that the CC-Expansion method does not seem to improve the object-level metrics much. We find two reasons for this. Firstly, not many instances are generated in this step, and the cells produced by this method are very small, which leads to minimal impact on DICE or AJI as those metrics are weighted based on the size of the instances. 
% \lgg{change this next part. talk about small cell performance}
Next, we observe that the Expand step improves small-cell AJI and DICE significantly in \textbf{all settings (highlighted in green)}, which validates its efficacy in identifying cells missed in the segmentation map.
% demonstrating the effectiveness of LRP in identifying cells missed in the segmentation maps. 
% Also, we find that the improvement in overall DICE and AJI is marginal in comparison, since most instances generated by LRP explanation are smaller cells.
Although the improvement of the Expand step on AJI and object-level DICE is not very significant, this is to be expected, as the Expand step targets small cells, which have minimal impact on those metrics.
% \reh{Upon visual inspection, we find that some small cells identified are not annotated properly, which can explain the less significant improvement in performance.}
% Secondly, upon a visual inspection, we find that many of the small cells detected by this method are not annotated properly. 
% An example is circled in blue in Figure \ref{testing_fig} and more examples can be seen in the supplementary material. 
% Usage of the CC-Expansion method seems more situational, and will be useful when there are small and obscure cells.
Qualitative visualization is provided in the Supplementary.

% Overall, our full \emph{Split} and \emph{Expand} method produces substantial gains \textbf{over all metrics and all settings}.
% It is worthwhile to note that this is achieved without explicit training in our \emph{Split} and \emph{Expand} method.

\begin{table}
\vspace{-0.3cm}
% \caption{FRW Results on MoNuSeg dataset. Best results in bold.}
\caption{Evaluation of FRW loss on MoNuSeg dataset. Best results are in bold.}
\vspace{-0.2cm}
\label{frw_tab}
\centering

\begin{tabular}{|l|l|p{0.12\textwidth}|p{0.12\textwidth}|p{0.16\textwidth}|p{0.16\textwidth}|}
\hline
Methods & Configuration & AJI & DICE & AJI (small) & DICE (small) \\
\hline
\multirow{3}{*}{\begin{tabular}{c@{}} Weakly\cite{qu2019_weakly} \end{tabular}} 
& Without FRW & 0.518 & 0.706 & 0.219 & 0.399 \\
& With FRW & \textbf{0.524} & \textbf{0.710} & \textbf{0.221} & \textbf{0.409} \\

\hline
\multirow{3}{*}{\begin{tabular}{c@{}} MaskGA\cite{guo2021learning} \end{tabular}} 
& Without FRW & 0.538 & 0.728 & 0.217 & 0.411 \\
& With FRW & \textbf{0.551} & \textbf{0.732} & \textbf{0.228} & \textbf{0.417} \\
\hline
\end{tabular}
\vspace{-1.5em}
\end{table}

Lastly, we evaluate the impact of the FRW loss in Table \ref{frw_tab} and find that it provides consistent improvements over all metrics, which shows its effectiveness.
\vspace{-1em}
\section{Conclusion}
This paper proposes a novel two-step inference-time method (\emph{Split} and \emph{Expand}) for weakly supervised cell instance segmentation that is \textit{training-free} and can be \textit{easily applied} to many existing baseline segmentation models. 
A novel FRW loss based on explanation methods is also proposed to help improve cell-center predictions.
% We train our model to also output CC predictions but no further explicit training is needed. 
% We test the method on MoNuSeg and TNBC datasets, observing both quantitative and qualitative improvements.
We test the method on MoNuSeg and TNBC datasets, observing significant object-level improvements.
\bibliographystyle{splncs04}
\bibliography{main}
%Author, F.: Article title. Journal \textbf{2}(5), 99--110 (2016)

%Author, F., Author, S.: Title of a proceedings paper. In: Editor,
%F., Editor, S. (eds.) CONFERENCE 2016, LNCS, vol. 9999, pp. 1--13.
%Springer, Heidelberg (2016). \doi{10.10007/1234567890}

%Author, F., Author, S., Author, T.: Book title. 2nd edn. Publisher,

%Author, A.-B.: Contribution title. In: 9th International Proceedings
%on Proceedings, pp. 1--2. Publisher, Location (2010)

%LNCS Homepage, \url{http://www.springer.com/lncs}. Last accessed 4
%\end{thebibliography}

\end{document}